\documentclass{article}

\usepackage[preprint]{neurips_data_2023}
\usepackage{amsmath,amsfonts,bm}

\def\eqref#1{equation~\ref{#1}}
\def\1{\bm{1}}

\def\rvu{{\mathbf{i}}}

\def\rvu{{\mathbf{u}}}

\def\rvx{{\mathbf{x}}}

\DeclareMathAlphabet{\mathsfit}{\encodingdefault}{\sfdefault}{m}{sl}
\SetMathAlphabet{\mathsfit}{bold}{\encodingdefault}{\sfdefault}{bx}{n}

\def\gG{{\mathcal{G}}}

\def\sM{{\mathbb{M}}}

\def\sU{{\mathbb{U}}}

\def\sX{{\mathbb{X}}}

\newcommand{\parents}{Pa} % See usage in notation.tex. Chosen to match Daphne's book.

\usepackage[utf8]{inputenc} % allow utf-8 input
\usepackage[T1]{fontenc}    % use 8-bit T1 fonts
\usepackage{hyperref}       % hyperlinks
\usepackage{url}            % simple URL typesetting
\usepackage{booktabs}       % professional-quality tables
\usepackage{amsfonts}       % blackboard math symbols
\usepackage{nicefrac}       % compact symbols for 1/2, etc.
\usepackage{microtype}      % microtypography
\usepackage{xcolor}         % colors
\usepackage{xr}
\usepackage{xcite}
\usepackage{hyperref}
\usepackage{cleveref}
\usepackage{enumitem}
\usepackage{subcaption}
\usepackage{url}
\usepackage{numprint}
\usepackage{booktabs}
\usepackage[pdftex]{graphicx}
\usepackage{listings}
\usepackage[normalem]{ulem}
\usepackage{chngcntr}
\usepackage{natbib}
\usepackage{color, colortbl}
\usepackage{xcolor}
\usepackage{adjustbox}

\definecolor{Col3}{gray}{0.92}
\definecolor{Col1}{gray}{0.88}
\definecolor{Col2}{RGB}{252, 243, 230}%{242, 222, 172}

\usepackage{tikz}
\usetikzlibrary{shapes.callouts}
\usepackage{float}
\newcommand{\themethod}{CausalBench}

\makeatletter
\newcommand*{\addFileDependency}[1]{% argument=file name and extension
\typeout{(#1)}% latexmk will find this if $recorder=0
\@addtofilelist{#1}
\IfFileExists{#1}{}{\typeout{No file #1.}}
}\makeatother

\newcommand*{\myexternaldocument}[1]{%
\externaldocument{#1}%
\addFileDependency{#1.tex}%
\addFileDependency{#1.aux}%
}

\myexternaldocument{supplementary_material}
\externalcitedocument{supplementary_material}
\title{\themethod{}: A Large-scale Benchmark for Network Inference from Single-cell Perturbation Data}

\author{%
Mathieu Chevalley$^{1,2}$ \quad Yusuf H Roohani$^{1,3}$ \quad Arash Mehrjou$^1$ \\
\textbf{Jure Leskovec}$^3$ \quad \textbf{Patrick Schwab}$^1$\\
$^1$GSK.ai \quad $^2$ETH Z\"urich \quad $^3$Stanford University\\
}

\begin{document}

\maketitle

\begin{abstract}
Causal inference is a vital aspect of multiple scientific disciplines and is routinely applied to high-impact applications such as medicine. However, evaluating the performance of causal inference methods in real-world environments is challenging due to the need for observations under both interventional and control conditions. Traditional evaluations conducted on synthetic datasets do not reflect the performance in real-world systems. To address this, we introduce \themethod{}, a benchmark suite for evaluating network inference methods on real-world interventional data from large-scale single-cell perturbation experiments. \themethod{} incorporates biologically-motivated performance metrics, including new distribution-based interventional metrics. 
A systematic evaluation of state-of-the-art causal inference methods using our CausalBench suite highlights how poor scalability of current methods limits performance.
Moreover, methods that use interventional information do not outperform those that only use observational data, contrary to what is observed on synthetic benchmarks. 
Thus, \themethod{} opens new avenues in causal network inference research and provides a principled and reliable way to track progress in leveraging real-world interventional data.
\end{abstract}

\section{Introduction}

\looseness -1 Causal inference is central to a number of disciplines including science, engineering, medicine and the social sciences. Causal inference methods are routinely applied to high-impact applications such as interpreting results from clinical trials \citep{farmer2018application}, studying the links between human behavior and economic activity \citep{baum2015causal}, optimizing complex engineering systems, and identifying optimal policy choices to enhance public health \citep{joffe2012causal}. However, evaluating these methods in real-world environments poses a significant challenge due to the time, cost, and ethical considerations associated with large-scale interventions under both interventional and control conditions. Consequently, most algorithmic development in the field has traditionally relied on synthetic datasets for evaluating causal inference approaches. Nevertheless, previous work \citep{gentzel2019case} has shown that such evaluations do not provide sufficient information on whether these methods generalize to real-world systems.

\looseness -1 Recent advancements in high-throughput biological experimentation have created real-world systems where large-scale interventions are feasible along with a directly observable comprehensive phenotypic readout (\cite{dixit2016perturb,datlinger2017pooled,datlinger2021ultra}). These datasets, which capture gene expression changes following interventions to individual cells, can serve as a valuable resource for evaluating the performance of observational and interventional approaches to causal inference on real-world data. Moreover, understanding the causal relationships between the expression of genes and constructing gene regulatory networks (GRNs) can also have a profound impact on therapeutic design and drug efficacy \cite{nelson2015support,yu2004advances,chai2014review,akers2021gene,hu2020integration}. However, effectively utilizing such datasets remains challenging, as establishing a causal ground truth for evaluating and comparing graphical network inference methods is difficult (\cite{neal2020realcause,shimoni2018benchmarking,parikh2022validating}). Furthermore, there is a need for systematic and well-validated benchmarks to objectively compare methods that aim to advance the causal interpretation of real-world interventional datasets while moving beyond reductionist (semi-)synthetic experiments.

\looseness -1 To facilitate the advancement of machine learning methods in this challenging domain, we introduce \themethod{} – the largest openly available benchmark suite for evaluating network inference methods on real-world interventional data. \themethod{} contains meaningful biologically-motivated performance metrics, a curated set of two large-scale perturbational single-cell RNA sequencing experiments with over \numprint{200000} interventional samples (each of which are openly available), and integrates numerous baseline implementations of state-of-the-art methods for causal network inference. Similar to other domains, e.g. ImageNet in computer vision (\cite{deng2009imagenet}), \themethod{} can accelerate progress on large-scale real-world causal graph inference. Our benchmarking results highlight how poor scalability and inadequate utilization of the interventional data limits performance. \themethod{} thus opens new research avenues and provides the necessary architecture to test future methodological developments. The source code is openly available at \url{https://github.com/causalbench/causalbench} under Apache 2.0 license.

\begin{figure}[t!]
    \centering
    \includegraphics[width=0.8\textwidth]{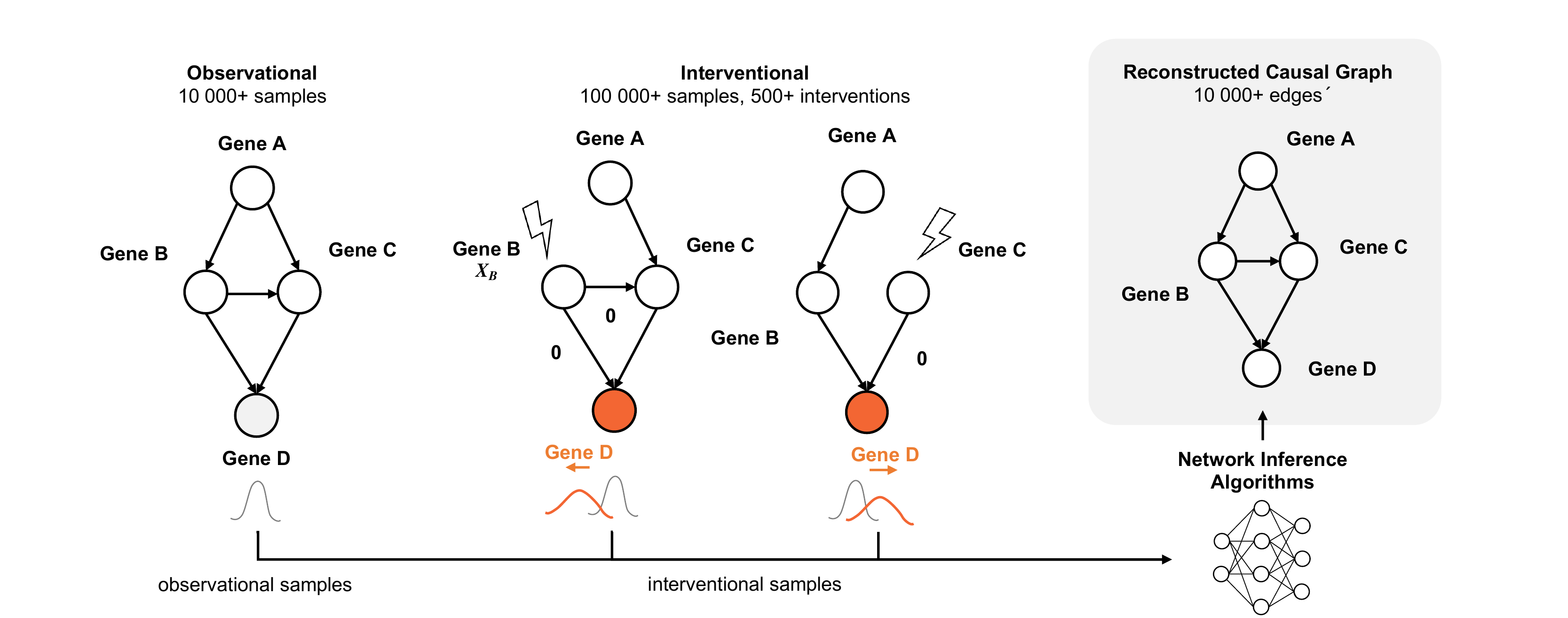}
    \caption{An overview of causal gene–gene network inference in mixed observational and perturbational single-cell data. The causal generative process in its unperturbed form is observed in the observational data (left; \numprint{10000}+ samples in \themethod{}) while data under genetic interventions (e.g. CRISPR knockouts) are observed in the interventional data (right; \numprint{200000}+ samples in \themethod{}). Either observational or interventional plus observational data that were sampled from the true causal generative process (bottom distributions) can be used by network inference algorithms (bottom right) to infer a reconstructed causal graph (top right) that should as closely as possible recapitulate the original underlying functional gene–gene interactions.
    }
    \label{fig:overview}
\end{figure}

\paragraph{Contributions.} \looseness -1  Our contributions are as follows:

\begin{itemize}[noitemsep, leftmargin=*]
\item We introduce \themethod{} – the most comprehensive benchmark suite for evaluating network inference methods using interventional data from over \numprint{400000} interventional samples.
\item  We introduce a set of meaningful benchmark metrics for evaluating performance, including novel statistical metrics that leverage single-cell perturbational data to evaluate the capability of test methods in recovering strong interventional effects and minimizing the rate of omission of causal relationships.
\item Using \themethod{}, we conduct a comprehensive experimental evaluation of the performance of state-of-the-art network inference algorithms. 
This includes an analysis of performance scaling characteristics under varying numbers of training samples and intervention set sizes.
\item Our results highlight an inconsistency between performance on synthetic data and real-world data. Specifically, we found that methods incorporating interventional information did not surpass observational methods, and causal methods did not outperform non-causal ones (methods not derived from causality theory). This finding underscores the necessity of benchmarks like \themethod{} for providing more accurate assessments of causal methods using real-world interventional data.
\end{itemize}

\section{Related Work}

\paragraph{Background.}\looseness -1 Given an observational data distribution, several different causal networks or directed acyclic graphs (DAGs) could be shown to have generated the data. The causal networks that could equally represent the generative process of an observational data distribution are collectively referred to as the Markov equivalence class (MEC) of that DAG (\cite{huang2018generalized}). Interventional data offer an important tool for limiting the size of the MEC to improve the identifiability of the true underlying causal DAG \citep{katz2019size}. In the case of gene expression data, modern gene-editing tools such as CRISPR offer a powerful mechanism for performing interventional experiments at scale by altering the expression of specific genes and observing the resulting interventional distribution across the entire transcriptome (\cite{dixit2016perturb,datlinger2017pooled,datlinger2021ultra}). The ability to leverage such interventional experiment data at scale could significantly improve our ability to uncover underlying causal functional relationships between genes and thereby strengthen our quantitative understanding of biology. Establishing causal links between genes can help implicate genes in biological processes causally involved in disease states and thereby open up new opportunities for therapeutic development (\cite{mehrjou2021genedisco,shifrut2018genome}). 

\paragraph{Network inference in mixed observational and interventional data.} \looseness -1 Learning network structure from both observational and interventional data presents significant potential in reducing the search space over all possible causal graphs. Traditionally, this network inference problem has been solved using discrete methods such as  permutation-based approaches (\cite{wang2017permutation, hauser2012characterization}. Recently, several new models have been proposed that can differentiably learn causal structure (\cite{scholkopf2021toward}). However, most of these models focus on observational datasets alone. \cite{ke2019learning} presented the first differentiable causal learning approach using both observational and interventional data. \cite{lopez2022large} improved the scalability of differentiable causal network discovery for large, high-dimensional datasets by using factor graphs to restrict the search space, and \citet{scherrer2021activecausal} introduced an active learning strategy for selecting interventions to optimize differentiable graph inference.

\paragraph{Gene regulatory network inference.} \looseness -1 The problem of GRN inference has been studied extensively in the bioinformatics literature in the case of observational datasets. Early work modeled this problem using a Bayesian network trained on bulk gene expression data (\cite{friedman2000using}). Subsequent papers approached this as a feature ranking problem where machine learning methods such as linear regression (\cite{kamimoto2020celloracle}) or random forests (\cite{huynh2010inferring, aibar2017scenic}) are used to predict the expression of any one gene using the expression of all other genes. However, \cite{pratapa2020benchmarking} showed that most GRN inference methods for observational data perform quite poorly when applied to single-cell datasets due to the large size and noisiness of the data. In the case of constructing networks using interventional data, there is relatively much less work given the recent development of this experimental technology. \cite{dixit2016perturb} were the first to apply network inference methods to single-cell interventional datasets using linear regression.

\paragraph{Benchmarks for causal discovery methods.} \looseness -1 To benchmark new causal discovery methods, the main approach followed consists of evaluating on purely synthetic data. The true underlying graph is usually drawn from a distribution of graphs, following procedures described in \citet{erdHos1960evolution} (Erdös-Rényi graphs) and \citet{barabasi1999emergence} (scale-free graphs). Given a drawn graph, a dataset is then created under an additive noise model assumption, following some functional relationship such as linear or nonlinear functions of random Fourier features. The additive noise can follow various distributions, such as Gaussian or uniform. The predicted graph is then evaluated using structural metrics that compare the prediction to the true graph. Popular metrics include precision, recall, F1, structural hamming distance (SHD) and structural intervention distance (SID) \citep{peters2015structural}. While this type of evaluation is valuable as a first validation of a new method in a controlled setting, it is limited in its capacity to predict the transportability of a method to a real-world setting. Indeed, the generated synthetic data tend to match the assumption of the proposed method, and the complexity of their distribution is reduced and uninformedly far from the distribution of real-world empirical data. 
To remedy this, synthetic data generators that mimic real-world data have been proposed, such as \citet{dibaeinia2020sergio} for the GRN domain. However, these simulated synthetic datasets are still limited and very reductionist, as they do not account for \emph{unknown-unknowns} in the true data-generating process. They also still offer a large set of degrees of freedom in the choice of hyperparameters ("researcher degrees of freedom" phenomenon \citep{simmons2011false}). Furthermore, the distribution over the true underlying graph still needs to be chosen, and how far those generated graphs are from a realistic causal graph is also unknown. As such, \citet{gentzel2019case} thoroughly argues that the evaluation of causal methods should incorporate evaluative mechanisms that examine empirical interventional metrics as opposed to purely structural ones, and that such evaluations are best performed using real empirical data.

\paragraph{Benchmarks for gene regulatory network inference}
\looseness -1 Past work has looked at benchmarking of different GRN inference approaches using single-cell gene expression data \cite{pratapa2020benchmarking}. However, this work only considers observational data and looks at small datasets of approximately 5000 samples (cells). Moreover, it does not benchmark most state-of-the-art causal inference methods. 

\section{Problem Formulation}

We introduce the framework of Structural Causal Models (SCMs) to serve as a causal language for describing methods, assumptions, and limitations and to motivate the quantitative metrics. The data's perturbational nature requires this formal statistical language beyond associations and correlations. We use the causal view as was introduced by \citet{pearl2009causality}.

\subsection{Structural Causal Models (SCMs).} Formally, an SCM $\sM$ consists of a 4-tuple $\left ( \sU, \sX, \mathcal{F}, P(\rvu) \right )$, where $\sU$ is a set of unobserved (latent) variables and $\sX$ is the set of observed (measured) variables \citep{peters2017elements}. $\mathcal{F}$ is a set of function such that for each $X_i \in \sX$, $X_i \leftarrow f_i(\parents_i, U_i)$, $U_i \in \sU$ and $\parents_i \in \sX \setminus X_i$. The SCM induces a distribution over the observed variables $P(\rvx)$. 
The variable-parent relationships can be represented in a directed graph, where each $X_i$ is a node in the graph, and there is a directed edge between all $\parents_i$ to $X_i$. The task of causal discovery can then be described as learning this graph over the variables. 
In the most general sense, an intervention on a variable $X_i$ can be thought as uniformly replacing its structural assignment with a new function $X_i \leftarrow \tilde{f}(X_{\widetilde{\parents}_i}, \tilde{U}_i)$. 
In this work, we consider the gene perturbation as being atomic or stochastic intervention and denote an intervention on $X_i$ as $\sigma(X_i)$. We can then describe the interventional distribution, denoted $P^{\sigma(X_i)}(\rvx)$, as the distribution entailed by the modified SCM. 
For consistency, we denote the observational distribution as both $P^{\emptyset}(\rvx)$ and $P(\rvx)$. This SCM framework is used throughout the paper.

\subsection{Problem Setting.} We consider the setting where we are given a dataset of vector samples $\rvx \in \mathbf{R}^d$, where $\rvx_i$ represents the measured expression of gene $i$ in a given cell.  
The goal of a graph inference method is to learn a causal graph $\gG$, where each node is a single gene. The causal graph $\gG$ induces a distribution over observed sample $P(\rvx)$, such that:
\begin{equation}
    P(\rvx) = \prod p(x_i | Pa_i)
\end{equation}

The datasets contain data sampled from $P^{\emptyset}(\rvx)$, as well as $P^{\sigma(X_i)}(\rvx)$ for various $i$. The observational setting only uses samples from $P^{\emptyset}(\rvx)$, the interventional setting includes observational and interventional data, while the partial interventional setting is a mix of the two, where only a subset of the genes are observed under perturbation. 

\subsection{Assumptions and challenges.} Single-cell data presents idiosyncratic challenges that may break the common assumptions of many existing methods. Apart from the high-dimensionality in terms of number of variables and large sample size, the distribution of the gene expression present a challenge as it is highly tailed at $0$ for some genes (\cite{tracy2019rescue}). Regarding the underlying true causal graph, biological feedback loops may break the acyclicity assumption underlying the use of DAGs. In addition, different cells sampled from the same batch may not be truly independent as the cells may have interacted and influenced their states. Lastly, cells in scRNAseq experiments are measured at a fixed point in time and may therefore have been sampled at various points in their developmental trajectory or cell cycle (\cite{kowalczyk2015single}) - making sampling time a potential confounding factor in any analysis of scRNAseq data.

\section{Benchmarking Setup}

\paragraph{Datasets for causal network inference}
\looseness -1 Effective GRN inference relies on gene expression datasets of sufficiently large size to infer underlying transcriptional relationships between genes. The size of the MECs inferred by these methods can hypothetically be reduced by using interventional data \citep{katz2019size}. For our analysis, we make use of gene expression data measured at the resolution of individual cells following a wide range of genetic perturbations. Each perturbation corresponds to the knock down of a specific gene using CRISPRi gene-editing technology \citep{larson2013crispr}. This is the largest and best quality dataset of its kind that is publicly available \citep{peidli2022scperturb}, and includes two different biological contexts (cell lines RPE-1 and K562). The dataset is provided in a standardized format that does not requires specialized libraries. We have also preprocessed the data so that the expression counts are normalized across batches and perturbations that appear to not have knocked down their target successfully are removed. The goal was to ensure that this benchmark is readily accessible to a broad machine learning audience while requiring no prior domain knowledge of biology beyond what is in this paper. A summary of the resulting two datasets can be found in \Cref{tab:dataset}. We hold out $20\%$ of the data for evaluation, stratified by intervention target.

\paragraph{Preprocessing} We incorporated a two-level quality control mechanism for processing our data, considering the perturbations and individual cells separately. 

In the perturbation-level control, we identified 'strong' perturbations based on three criteria adopted from \citet{replogle2022mapping}: (1) inducing at least 50 differentially expressed genes with a significance of p < 0.05 according to the Anderson-Darling test after Benjamini-Hochberg correction; (2) being represented in a minimum of 25 cells passing our quality filters; and (3) achieving an on-target knockdown of at least 30\%, if measured. 

In the individual cell-level control, we checked the effectiveness of each perturbation by contrasting the expression level of the perturbed gene (X) post-perturbation against its baseline level. We established a threshold for expression level at the 10th percentile of gene X in the unperturbed control distribution. We excluded any cell where gene X was perturbed but its expression exceeded this threshold. However, if a perturbed gene's expression was not measured in the dataset, we did not perform this filtering.

\begin{table}[t!]
    \caption{High-level description of the two large-scale datasets utilized in \themethod{} – characterized by high numbers of samples and intervened-on variables. Of those samples, after stratification by intervention target (including no target), $20\%$ were kept as held-out data for evaluation.}
    \centering
    \begin{tabular}{p{11.8em}|rrr}
        \toprule
        Dataset & Total Samples & \# Observational Samples & \# Gene Interventions \\
        \midrule
         \citep{replogle2022mapping} K562 & \numprint{162751} & \numprint{10691} & \numprint{622} \\
         \citep{replogle2022mapping} RPE1 & \numprint{162733} & \numprint{11485} & \numprint{383} \\
         \bottomrule
    \end{tabular}
    \label{tab:dataset}
\end{table}

\paragraph{Network inference model input and output} \looseness -1 The benchmarked methods are given either observational data only or both observational and interventional data – depending on the setting – consisting of the expression of each gene in each cell. For interventional data, the target gene in each cell is also given as input. We do not enforce that the methods need to learn a graph on all the variables, and further preprocessing and variables selection is permitted. The only expected output is a list of gene pairs that represent directed edges. No properties of the output network, such as acyclicity, are enforced either. 

\paragraph{Baseline models} \looseness -1 We implement a representative set of existing state-of-the-art methods for the task of causal discovery from single-cell observational and mixed perturbational data. For the observational setting, we implement PC (named after the inventors, Peter and Clark; a constraint-based method) \citep{spirtes2000causation}, Greedy Equivalence Search (GES; a score-based method) \citep{chickering2002optimal}, and NOTEARS variants NOTEARS (Linear), NOTEARS (Linear, L1), NOTEARS (MLP) and NOTEARS (MLP, L1) \citep{zheng2018dags, zheng2020learning}, Sortnregress \citep{reisach2021beware} (a marginal variance based method) and GRNBoost \citep{aibar2017scenic} (a tree-based GRN inference method). In the interventional setting, we included Greedy Interventional Equivalence Search (GIES, a score-based method and extension to GES) \citep{hauser2012characterization}, the Differentiable Causal Discovery from Interventional Data (DCDI) variants DCDI-G and DCDI-DSF (continuous optimization-based methods) \citep{brouillard2020differentiable}, and DCDI-FG \citep{lopez2022large}.  GES and GIES greedily add and remove edges until a score on the graph is maximized. NOTEARS, DCDI-FG, and DCDI enforce acyclicity via a continuously differentiable constraint, making them suitable for deep learning. 
\vspace{-1em}
\paragraph{Benchmark usage} \looseness -1 \themethod{} has been designed to be easy to setup and use. Adding or testing a new method is also straightfoward. A more comprehensive guide can be found in the README of the Github repository (\url{https://github.com/causalbench/causalbench}) as well as the starter repository of the \themethod{} challenge (\url{https://github.com/causalbench/causalbench-starter/}).

\section{Evaluation}

\looseness -1 In \themethod{}, unlike standard benchmarks with known or simulated graphs, the true causal graph is unknown due to the complex biological processes involved. This absence of a definitive ground truth presents unique challenges when trying to evidence the superiority of one method over another. We respond to this by developing synergistic metrics with the aim of measuring the accuracy of the output network in representing the underlying complex biological processes. We employ two evaluation types: a biology-driven approximation of ground truth and a quantitative statistical evaluation. The former, though not fully cell-specific, adds an extra validation layer based on known biology. However, we mainly use a new statistical metric for model comparison, which is independent of prior knowledge and data-dependent, and correlates well with relevant downstream applications.

\subsection{Biological evaluation}

\looseness -1 The biologically-motivated evaluation in \themethod{} is based on biological databases of known putatively causal gene–gene interactions. Using these databases of domain knowledge, we can construct putatively true undirected subnetworks to evaluate the output networks in the understanding that the discovered edges that are not present in those databases are not necessarily false positives. We can then compute metrics such as precision and recall against those constructed networks. 

\paragraph{Biological evaluations metrics} 
To implement the biologically-motivated evaluation, we extract network data from two widely used open biological databases: CORUM \citep{giurgiu2019corum} and STRING \citep{von2005string, snel2000string, von2007string, jensen2009string, mering2003string, szklarczyk2010string, szklarczyk2015string, szklarczyk2016string, szklarczyk2019string, szklarczyk2021string, franceschini2012string, franceschini2016svd}. CORUM is a repository of experimentally characterized protein complexes from mammalian organisms. The complexes are extracted from individual experimental publications, and exclude 
results from high-throughput experiments. We extract the human protein complexes from the CORUM repository and aggregate them to form a network of genes. STRING is a repository of known and predicted protein-protein interactions. STRING contains both physical (direct) interactions and indirect (functional) interactions that we use to create two evaluation networks from STRING: Protein-protein interactions (network) and protein-protein interactions (physical).
Protein-protein interactions (physical) contains only physical interactions, whereas protein-protein interactions (network) contains all types of know and predicted interactions. STRING, and in particular string-network, can contain less reliable links, as the content of the database is pulled from a variety of evidence, such as high-throughput lab experiments, (conserved) co-expression, and text-mining of the literature. Recognizing the importance of considering cell type-specific effects when evaluating gene regulatory network inference methods, we also incorporate cell-type-specific networks for the biological evaluation. For the K562 cell line, we employed data from the Chip-Atlas \citep{zou2022chip} and ENCODE \citep{davis2018encyclopedia} databases to build a ChiPSeq networks, restricting the links to those relevant to or recorded in the K562 cell line. However, the RPE1 cell line has not been as comprehensively characterized in existing research. To address this, we utilized networks derived in a similar manner, but from a more extensively studied epithelial cell line, HepG2, given that RPE1 is also an epithelial cell line.

\subsection{Statistical evaluation}
\label{ssec:stat_eval}
\looseness -1 In contrast to the biologically-motivated evaluation, the statistical evaluation in \themethod{} is data-driven, cell-specific, and does not rely on prior knowledge. This evaluation method is uniquely designed for single-cell perturbational data and provides a new way to approximate ground-truth gene regulatory interactions, supplementing information found in biological databases. The evaluation uses the interventional data from perturbational scRNA-seq experiments to assess predicted edges in the output networks. Here, we thus closely follow the postulate of \citet{gentzel2019case} that causal methods should be evaluated on empirical data using interventional metrics that correlate with the strength of the underlying relationships.

\looseness -1 The main assumption for this evaluation is that if the predicted edge from $A$ to $B$ is a true edge denoting a functional interaction between the two genes, then perturbing gene $A$ should have a statistically significant effect on the distribution $P^{\sigma(X_A)}(\rvx_B)$ of values that gene $B$ takes in the transcription profile, compared to its observational distribution $P^{\emptyset}(\rvx_B)$ (i.e compared to control samples where no gene was perturbed).  Conversely, we can test for the predicted absence of gene interactions. We call a gene interaction $A$ to $B$ negative if there is no path in the predicted graph from $A$ to $B$. If no interaction exists in the true graph, then there should be no statistically significant change in the distribution of $B$ when intervening on $A$. We thus aim to estimate the false omission rate (FOR) of the predicted graph. The FOR is defined as:

\begin{equation*}
    \text{FOR} = \frac{\text{False Negatives}}{\text{False Negatives} + \text{True Negatives}}
\end{equation*}

\looseness -1 To test the predicted interactions, we propose using the mean Wasserstein distance. For each edge from $A$ to $B$, we compute a Wasserstein distance \citep{ramdas2017wasserstein} between the two empirical distributions of $P^{\sigma(X_A)}(\rvx_B)$ and $P^{\emptyset}(\rvx_B)$. We then return the mean Wasserstein distance of all inferred edges. We call this metric the mean Wasserstein. A higher mean Wasserstein distance should indicate a stronger interventional effect of intervening on the parent. 
Although the quantitative statistical approach cannot differentiate between causal effects from direct edges or causal paths in the graph, we expect direct relationships to have stronger causal effects \citep{meinshausen2016methods}. Despite this limitation, the quantitative evaluation offers a data-driven, cell-specific, and prior-free metric. It also correlates with the strength of the causal effects, and thus evaluates methods in their ability to recall the strongest perturbational effects, which is a downstream task of high interest. Moreover, it presents a novel approach for estimating ground-truth gene regulatory interactions that is uniquely made possible through the size and interventional nature of single-cell perturbational datasets. Then, we evaluate the predicted negative interactions of the output. To do so, we sample random pairs of genes such that there is no path in the predicted graph between the two. We then perform a two-sided Mann–Whitney U rank test \citep{mann1947test, Bucchianico1999-ul} between samples from $P^{\sigma(X_A)}(\rvx_B)$ and $P^{\emptyset}(\rvx_B)$ for all sampled negative pairs using the SciPy package \citep{2020SciPy-NMeth} to test the null hyptothesis that the two distributions are equal.  A rejected test, with a p-value threshold of $5\%$, indicates a false negative. The trade-off between maximizing the mean Wasserstein and minimizing the FOR exhibits the ranking nature of this applied task, as opposed to predicting against a fixed binary ground-truth. Contrary to structural metrics, errors in prediction are weighted given their causal importance \citep{gentzel2019case}. To our knowledge, we are also the first to propose evaluation of the negative predictions.

\begin{figure}[htp]
    \vspace{-1.5em}
    \centering
    \begin{adjustbox}{center, max width=11cm}
    \includegraphics[]{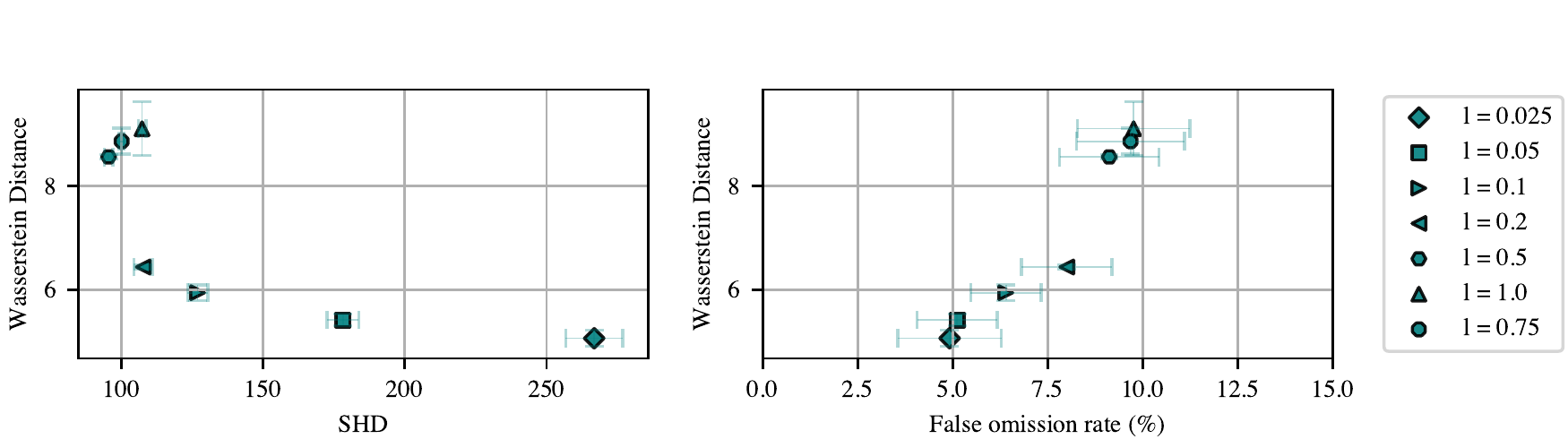}
    \end{adjustbox}
    \caption{Plots showing the characteristics of the proposed statistical evaluations, here validated on synthetic data. A NOTEARS (MLP, L1) model is run with different regularization values. Each setting is run five times, with the mean and standard deviation score plotted across the five runs for each setting.}
    \label{fig:synthetic_metrics}
\end{figure}

\paragraph{Analysis of the proposed quantitative evaluations}
\looseness -1 We here perform an analysis of the proposed metrics to validate their meaningfulness and to study their properties. We follow a procedure similar to \citet{gentzel2019case}. We create a synthetic dataset using a additive noise model with random Fourier features, reusing code from \citet{lorch2022amortized}. The training set consists of $500$ samples and the test set consists of $1500$ observational samples and $30$ interventional samples per variable, which makes the test set comparable to the test sets in \themethod{}. We then train the Notears (MLP, L1) model with various strengths of sparsity regularization ($l \in \{0.025, 0.05, 0.1, 0.2, 0.5, 0.75, 1.0\}$, repeated five times per value of $l$. The results are presented in \cref{fig:synthetic_metrics}. As can be observed, our proposed statistical metrics highlight the trade-off between recovering and omitting causal relationships, where a stronger regularization value leads to a smaller graph recalling the strongest causal relationship, but omitting many others. At the other end of the spectrum, we can see that at a value of $l = 0.05$ all causal relationships are recalled as the FOR is $5\%$, which is equal to the p-value threshold. 
Lastly, we can observe that the mean Wasserstein is well correlated with a structural metric such as SHD, but that it gives a better ordering of the models that is weighted in terms of strength of the predicted causal relationship. As such, our proposed statistical metrics are well suited for applied tasks and offer a meaningful tool for model comparison and selection in practice.

\subsection{Characteristics of the optimal method}
\label{ssec:charac} \looseness -1 For actual applicability to this real-world setting, the optimal method should exhibit properties that go beyond best relative performance on the metrics presented previously. First, to uncover interactions among all encoding genes, the optimal graph inference method should computationally scale to graphs with a large number of nodes (in the thousands). Furthermore, the method performance should increase as more samples and more targeted genes are given as input. This ensures that the method is future-proof as we expect larger datasets that target almost all possible genes to become ubiquitous in the near future. The scale of the datasets used here allows us to test for these scaling properties in our benchmark by making the creation of settings with varying fractions of data or of number of interventions easy and principled. 

\section{Benchmarking Results}

\begin{figure}[!t]
    \centering 
    \begin{adjustbox}{center, max width=11cm}
    \includegraphics{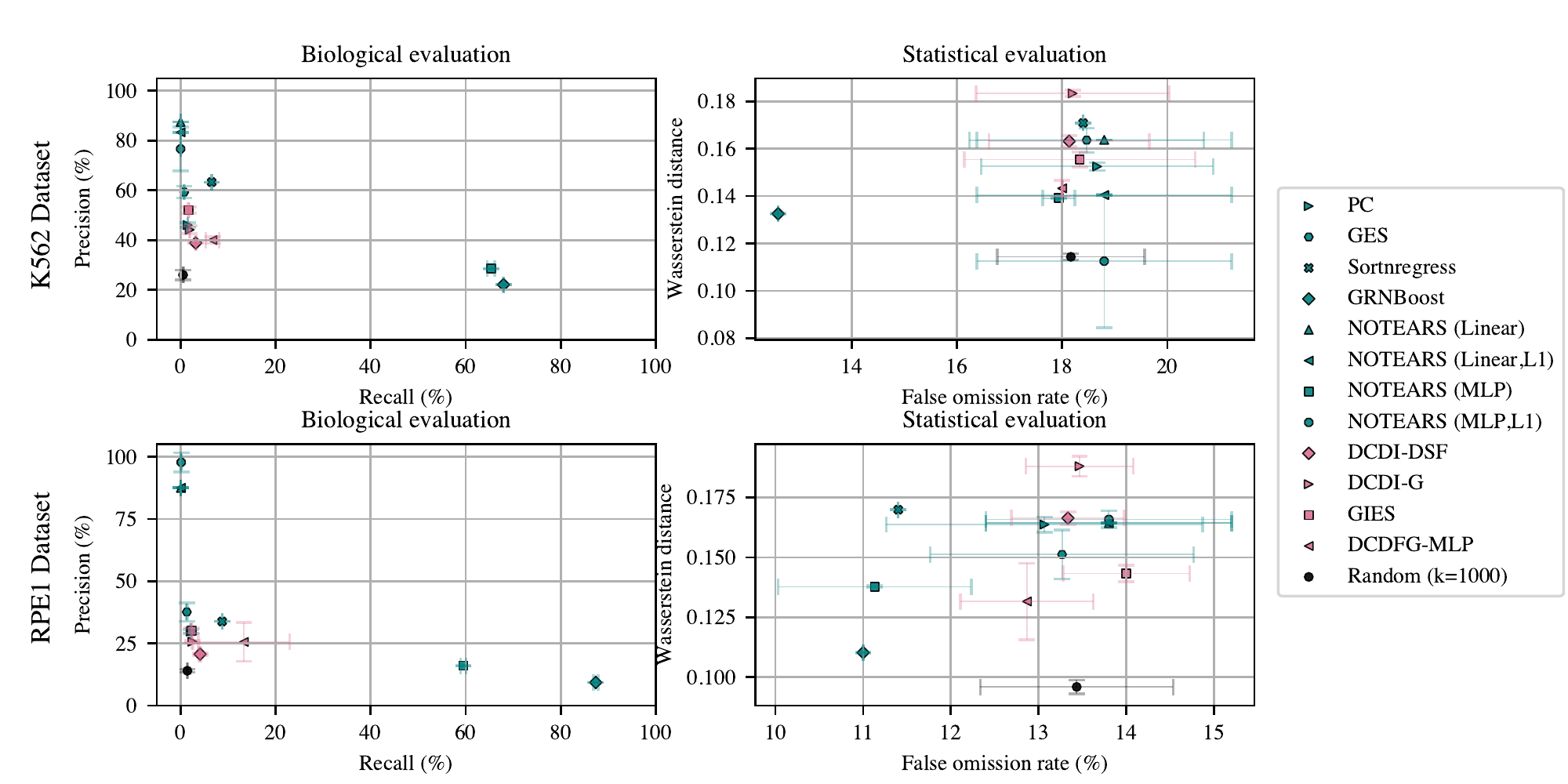}
    \end{adjustbox}
    \caption{Performance comparison in terms of Precision (in \%; y-axis) and Recall (in \%; x-axis) in correctly identifying edges substantiated by biological interaction databases (left panels); and our own statistical evaluation using interventional information in terms of Wasserstein distance and FOR (right panels). Performance is compared across 8 different methods using observational data (green markers), 4 different methods using interventional data (pink markers), and 1 random baseline (black markers) in K562 (top panels) and RPE1 (bottom panels) cell lines. For each method, we show the mean and standard deviation from three independent runs. Complete and detailed results can be found in \cref{app:results}.}
    \label{fig:precision_recall}
\end{figure}

\subsection{Network inference} 
\looseness -1 We here summarize the experimental results of our baselines in both observational and interventional settings, on the statistical evaluations and the biologically-motivated evaluations. All results are obtained by training on the full dataset three times with different random seeds. 

\paragraph{Trade-off between precision and recall}
\looseness -1 We highlight the trade-off between recall and precision. Indeed, we expect methods to optimize for these two goals, as we want to obtain a high precision while maximizing the percentage of discovered interactions. On both the biological and statistical evaluation, we observe this trade-off, as most of the baseline methods tend to gravitate toward one extreme or the other. That is, they either prioritize precision at the cost of uncovering fewer interactions, or they uncover a larger number of interactions but with less precision. GRNBoost is the only method with a high recall, but this comes with low precision. Most of the other baselines have similar recall and varying precision. Results are summarized in \cref{fig:precision_recall}. As can be observed, there is no major difference in performance between observational and interventional methods.

\paragraph{Causal models that leverage interventional information do not outperform the others}
\looseness -1 We observe that, contrary to what would be theoretically expected, interventional methods do not outperform observational methods, even though they are trained on more informative data. For example, on both datasets, GIES does not outperform its observational counterpart GES. Also, simple non-causal methods such as GRNBoost and Sortnregress perform highly. GRNBoost performance is mainly driven by its capacity to recall most causal interactions, at the expense of its precision. Sortnregress performance shows that even the varsortability \citep{reisach2021beware} of the two datasets is underexploited by most methods. Lastly, DCDI-G shows high mean Wasserstein, which demonstrates some capacity to leverage the interventional data, but it omits a large number of interactions given it cannot infer the graph on all variables. This highlights an opportunity for further method development for causal graph inference in order to be able to fully leverage interventional data. Early results in this direction were made possible using \themethod{} to organize a community machine learning challenge. To summarize the performance of the baselines, we propose a simple and unbiased way to compute a ranked scoreboard that takes into account the Wasserstein score and FOR. The detailed ranking can be found in \cref{app:ranking}, and a summary is given in \cref{tab:network_inference_models}.

\subsection{No model fulfills all the characteristics of an optimal method for this task}
\label{ssec:charac_analysis}

\paragraph{Most methods do not scale to the full graph.} \looseness -1 Unfortunately, all tested methods, with the exception of NOTEARS, GRNBoost, DCDFG and Sortnregress, do not computationally scale to graphs of the size typically encountered in transcriptome-wide GRN inference. Nonetheless, to enable a meaningful comparison, we partitioned the variables into smaller subsets where necessary and ran the methods on each subset independently, with the final output network being the union of the subnetworks. 
The proposed approach breaks the no-latent-confounder assumption that some methods may make, and it also does not fully leverage the information potentially available within the data. Methods that do scale to the full graph in a single optimization loop should therefore perform better.

\paragraph{Performance as a function of sample size.} \looseness -1 We additionally studied the effect of sample size on the performance of the evaluated state-of-the-art models. We randomly subset the data at different fractions and report the mean Wasserstein distance as explained in the quantitative evaluations part of Section \ref{ssec:stat_eval} and shown in \Cref{fig:wasserstein_distance_effect}. In the observational setting, the sample size does not seem to have a significant effect on performance for most methods, indeed having a slightly negative effect for some methods. 
In the interventional setting, a positive impact is observable for a larger sample size, especially for the methods that rely on deep networks and gradient-based learning such as DCDI, whereas GIES seems to suffer in a large sample setting. 

\paragraph{Performance by the fraction of perturbations.}  \looseness -1 Beyond the size of the training set, we also studied the partial interventional setting – where only a subset of the possible  genes to perturb are experimentally targeted. We adapt the fraction of randomly targeted genes from  5\% (low ratio of interventions) to 100\% (fully interventional). We randomly subset the genes at different fractions, using three different random seeds for each method, and report the mean Wasserstein distance as a measure of quantitative evaluation. We would expect a larger fraction of intervened genes to lead to higher performance, as this should facilitate the identification of the true causal graph. We indeed observe a better performance for DCDI on K562 as we increase the fraction of intervened genes. For GIES, we again observe a negative effect of having more data on RPE1. 

\begin{table}[]
\small
    \caption{Table summarizing the characteristics of the test baselines as laid out in \cref{ssec:charac} and analysed in \cref{ssec:charac_analysis}, as well as two per dataset rankings. Graph scaling refers to the ability to infer on the complete set of genes. A method is deemed to have successfully achieved sample scaling and intervention scaling if its performance improves by at least 10\% on both datasets. This comparison is made between settings where only a subset, specifically 25\% of samples or perturbations, is used and settings where 100\% of them are utilized. As such, no baseline fulfills all the criteria. Details on the ranking can be found in \cref{app:ranking}.}
    \centering
    \begin{tabular}{l|ccc}
        \toprule
         Model & Graph & Sample & Intervention\\
         &scaling & scaling & scaling\\
         \midrule
         PC & \texttimes & \texttimes & -- \\
         GES & \texttimes & \texttimes & --\\
         GIES & \texttimes & \texttimes & \texttimes\\
         %GSP && &\\
         %IGSP & \checkmark & \texttimes & \texttimes\\
         NOTEARS & \checkmark & \texttimes & --\\
         DCDI-G & \texttimes & \checkmark & \texttimes\\
         DCDI-DSF & \texttimes & \texttimes & \texttimes\\
         DCDI-FG & \checkmark & \texttimes & \texttimes\\
         GRNBoost & \checkmark & \texttimes & --\\
         Sortnregress & \checkmark & \texttimes & -- \\
         \bottomrule 
    \end{tabular}
    \begin{tabular}{l}
        \toprule
         Ranking K562  \\
        \midrule
         GRNBoost \\
             DCDI-G \\
       Sortnregress \\
           DCDI-DSF \\
                GES \\
   NOTEARS (Linear) \\
               GIES \\
                 PC \\
      NOTEARS (MLP) \\
          DCDFG-MLP \\
NOTEARS (Linear,L1) \\
   NOTEARS (MLP,L1) \\
         \bottomrule
    \end{tabular}
    \begin{tabular}{l}
        \toprule
         Ranking RPE1  \\
        \midrule
         Sortnregress \\
           GRNBoost \\
             DCDI-G \\
      NOTEARS (MLP) \\
                 PC \\
           DCDI-DSF \\
   NOTEARS (MLP,L1) \\
   NOTEARS (Linear) \\
NOTEARS (Linear,L1) \\
          DCDFG-MLP \\
                GES \\
               GIES \\
         \bottomrule
    \end{tabular}
    \label{tab:network_inference_models}
\end{table}

\subsection{Compute resources}
\label{ssec:resources}
\looseness -1 All methods were given the same computational resources, which consists of 20 CPU's with 32GB of memory each. We additionally assign a GPU for the DCDI and DCDFG methods. The hyperparameters of each method, such as partition sizes, are chosen such that the running time remains below $30$ hours. Partition sizes for each model can be found in \cref{app:partition}. 

\subsection{Limitations.} 
\label{ssec:limitations}
\looseness -1 Openly available benchmarks for causal models for large-scale single-cell data could potentially accelerate the development of new and effective approaches for uncovering gene regulatory relationships. However, some limitations to this approach remain: firstly, the biological networks used for evaluation do not fully capture ground-truth GRNs, and the reported connection are often biased towards better-studied systems and pathways \citep{gillis2014bias}. True ground-truth validation would require prospective and exhaustive interventional wet-lab experiments. However, at present, experiments at the scale necessary to exhaustively map gene–gene interactions across the genome are cost prohibitive for all possible edges. 
Beyond limitations in data sources used, there are limitations with some of the assumptions in the utilized state-of-the-art models. For instance, feedback loops between genes are a well-known phenomenon in gene regulation \citep{carthew2006gene,levine2005gene} that unfortunately cannot be represented by most existing causal network inference methods at present. While many causal discovery methods assume causal sufficiency, which posits that all common causes of any pair of variables are observed, it is important to note that this assumption may not hold true in the datasets included in \themethod{}. Lastly, the assumption of linearity of interaction may not hold for gene-gene interactions. We acknowledge these limitations and encourage ongoing research to address them.

\section{Conclusion}
\looseness -1 We introduced \themethod{} – a comprehensive open benchmark for evaluating causal discovery algorithms on real-world interventional data, using two large-scale CRISPR-based interventional scRNA-seq datasets. \themethod{} introduces a set of biologically meaningful performance metrics to quantitatively compare graphs that are proposed by causal inference methods using novel statistical evaluations, and validated them according to existing biological knowledge bases.  
\themethod{}  was designed to streamline the evaluation of causal discovery methods on real-world interventional data by standardizing non-model-related components of the evaluation process, thereby allowing researchers to focus on advancing causal network discovery methods. With more than \numprint{200000} interventional samples, \themethod{} is built on one of the largest open real-world interventional datasets \citep{replogle2022mapping}, providing the causal machine learning community with unprecedented access to large-scale interventional data for developing and evaluating causal discovery methods. Our results highlight many challenges faced by existing models applied to this challenging domain, such as poor scalability and inadequate utilization of the interventional data. Major strides in this direction have already been made, as evidenced by a recent machine learning community challenge that successfully utilized \themethod{} (see \url{https://www.gsk.ai/causalbench-challenge/} for details). 

\section*{Acknowledgements and disclosure}

The authors thank Prof. Nicolai Meinshausen for feedback and comments on the statistical evaluations and on the manuscript. We also thank Siobhan Sanford and Djordje Miladinovic for comments and edits on the manuscript. MC, YR, AM and PS are employees and shareholders of GSK plc.

\bibliography{reference}
\bibliographystyle{abbrvnat}

\appendix

\section{Baselines descriptions}

\subsubsection{PC} is one of the most widely used methods in causal inference from observational data that assumes there are no confounders and calculates conditional independence to give asymptotically correct results. It outputs the equivalence class of graphs that conform with the results of the conditional independence tests.

\subsubsection{Greedy Equivalence Search (GES)} implements a two-phase procedure (Forward and Backward phases that adds and removes edges from the graph) to calculate a score to choose within an equivalence class. While GES leverages only observational data, its extension, Greedy Interventional Equivalence Search (GIES), enhances GES by adding a turning phase to allow for the inclusion of interventional data. 

\subsubsection{NOTEARS} formulates the DAG inference problem as a continuous optimization over real-valued matrices that avoids the combinatorial search over acyclic graphs. This is achieved by constructing a smooth function with computable derivatives over the adjacent matrices that vanishes only when the associated graph is acyclic. Various versions of NOTEARS refer to which function approximator is employed (either a MLP or Linear) or which regularity term is added to the loss function (e.g. L1 for the sparsity constraint.)

\subsubsection{Differentiable Causal Discovery from Interventional Data (DCDI)} \citep{brouillard2020differentiable} leverages various types of interventions (perfect, imperfect, unknown), and uses a neural network model to capture conditional densities. DCDI encodes the DAG using a binary adjacency matrix. The intervention matrix is also modeled as a binary mask that determines which nodes are the target of intervention. A likelihood-based differentiable objective function is formed by using this parameterization, and subsequently maximized by gradient-based methods to infer the underlying DAG. DCDI-G assumes Gaussian conditional distributions while DCDI-DSF lifts this assumption by using normalizing flows to capture flexible distributions.

\subsubsection{GRNBoost} \citep{aibar2017scenic} is a GRN specific Gradient Boosting tree method, where for every gene, candidate parent gene are ranked based on their predictive power toward the expression profile of the downstream gene. As such, it acts as a feature selection method toward learning the graph. GRNBoost was identified as one of the best performing GRN method in previous observational data based benchmarks \citep{pratapa2020benchmarking}.

\subsubsection{Random (k)} is the simplest baseline which outputs a graph from which $k$ nodes are selected at uniformly random without replacement. In the experiments, we tested $k=100$, $1000$, and $10000$.

\section{Additional results}
\label{app:results}

\begin{figure}[!t]
    \centering 
    \begin{adjustbox}{center, max width=11cm}
    \includegraphics{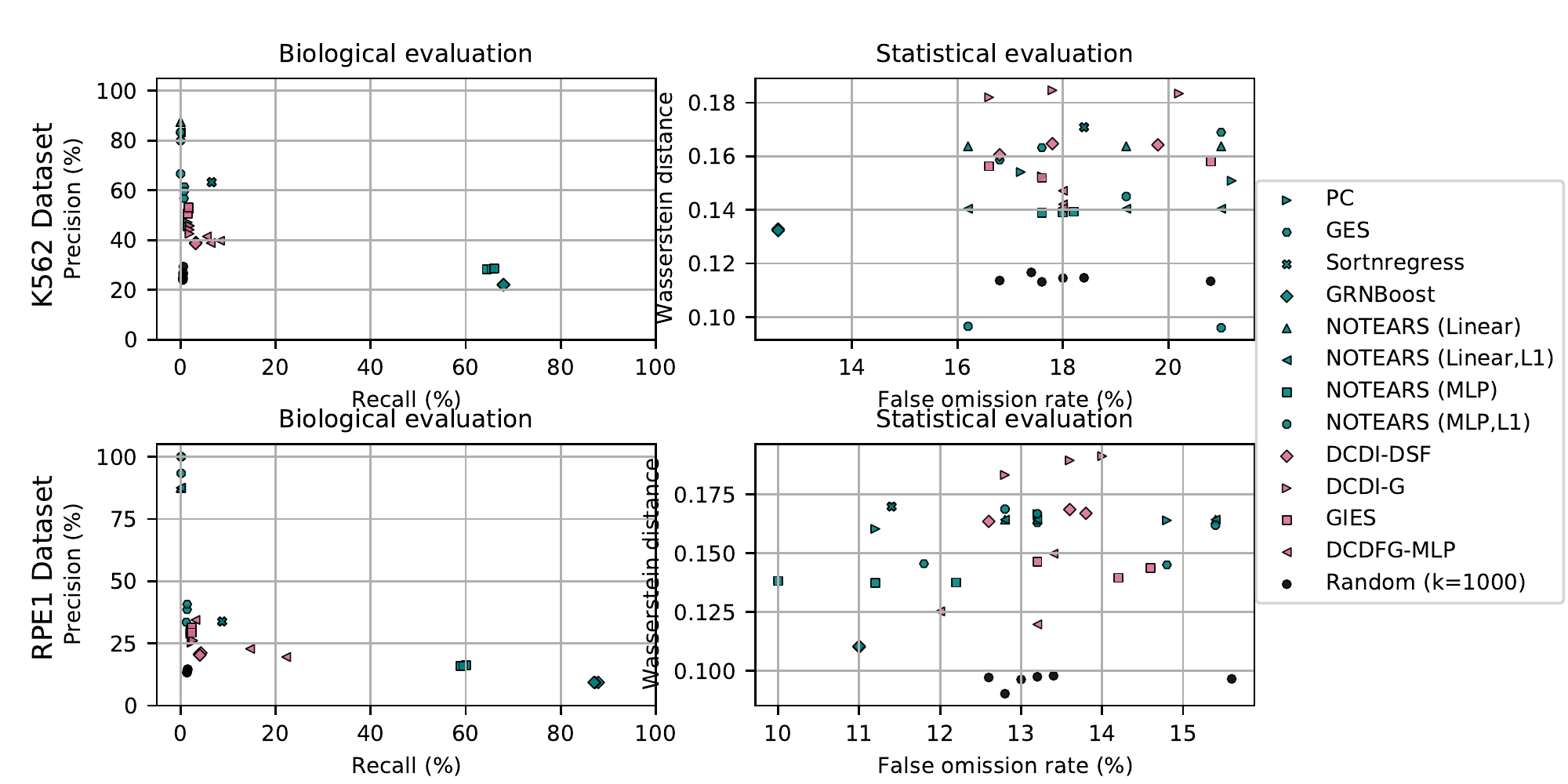}
    \end{adjustbox}
    \caption{Performance comparison in terms of Precision (in \%; y-axis) and Recall (in \%; x-axis) in correctly identifying edges substantiated by biological interaction databases (left panels); and our own statistical evaluation using interventional information in terms of Wasserstein distance and FOR (right panels). Performance is compared across 8 different methods using observational data (green markers), 4 different methods using interventional data (pink markers), and 1 random baseline (black markers) in K562 (top panels) and RPE1 (bottom panels) cell lines. For each method, we show the score of three independent runs. }
    \label{fig:precision_recall_app}
\end{figure}

\begin{figure}[]
    \vspace{-3em}
    \centering
        \centering
        \includegraphics[width=1.0\textwidth]{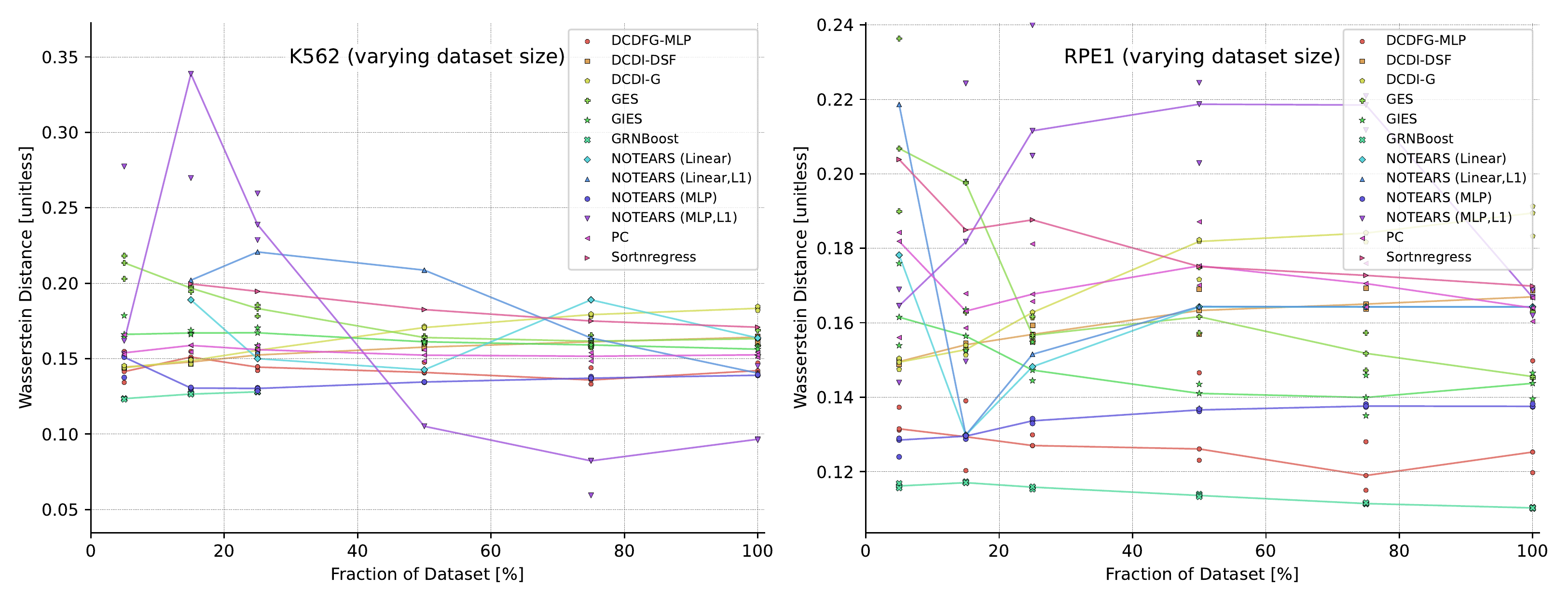}
        \includegraphics[width=1.0\textwidth]{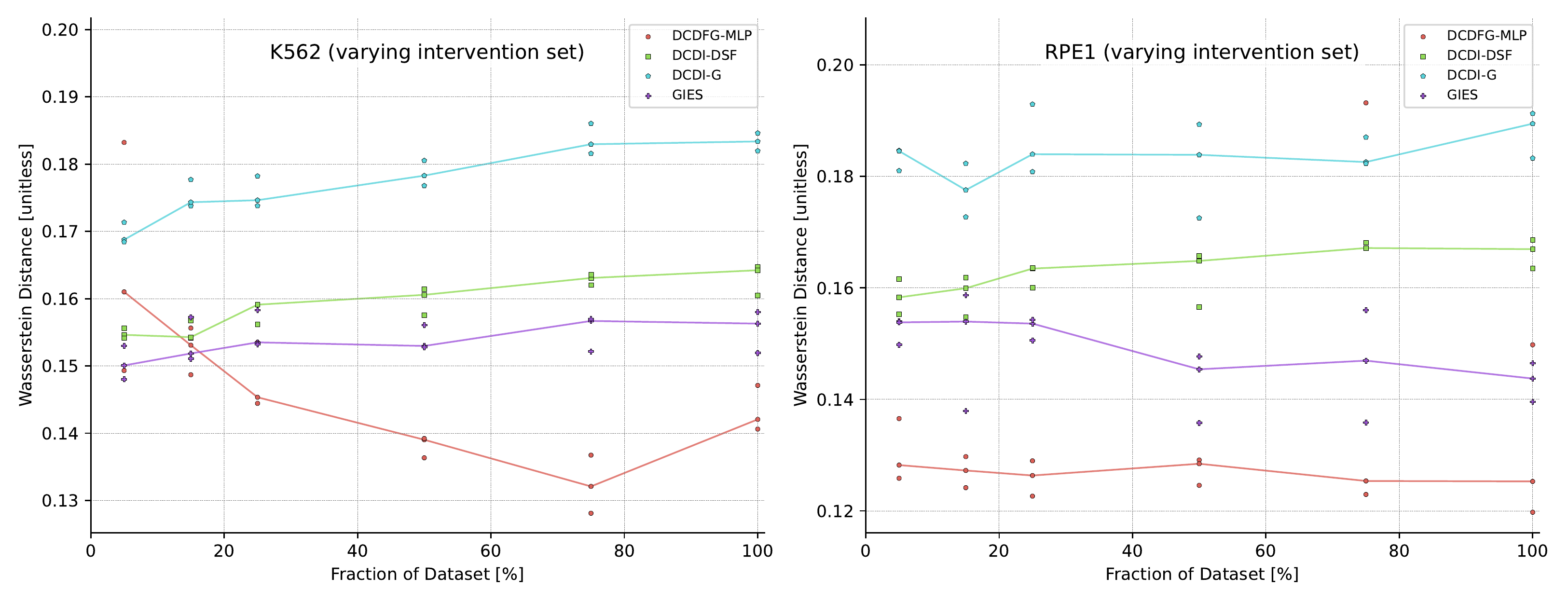}
    \caption{Performance comparison in terms of Mean Wasserstein Distance (unitless; y-axis) of 7 methods for causal graph inference on observational data (top row; (see legend top right) and 6 methods on interventional scRNAseq data (centre row; see legend centre right) when varying the fraction of the full dataset size available for inference (in \%; x-axis), and 6 methods on interventional data (bottom row; see legend bottom right) when varying the fraction of the full intervention set used (in \%, x-axis). Markers indicate the values observed when running the respective algorithms with one of three random seeds, and colored lines indicate the median value observed across all tested random seeds for a method.
    \label{fig:wasserstein_distance_effect}
    }
    
    \vspace{-2em}
\end{figure}

We here recapitulate more detailed and extensive results of our analysis of stat-of-the-art method using Causalbench. \Cref{fig:precision_recall_app} shows the same plot as in the main text but with single run as individual points. \Cref{tab:benchmark_k562} and \cref{tab:benchmark_rpe1} show the precision and recall scores for the biological evaluation for each database. \Cref{fig:wasserstein_distance_effect} shows the effect of varying the dataset or intervention set size for each method.

\begin{table*}[h!]
\centering
\caption{Performance results on RPE-1 Dataset for the biological evaluation, for each evidence type. Pooled correspond to pooling all the extracted databases into one network for evaluation.}
\footnotesize
\begin{adjustbox}{center, max width=11cm}
{\renewcommand{\arraystretch}{1.15}
\begin{tabular}{l|ccccc|ccccc}
\hline
Model&\multicolumn{5}{@{}c@{}|}{Precision}&\multicolumn{5}{@{}c@{}}{Recall}\\
\hline
&Pooled&CORUM&PPI (N)&PPI (P)&CHIP&Pooled&CORUM&PPI (N)&PPI (P)&CHIP\\
\hline

\rowcolor{Col1}[.22\tabcolsep][2.7mm]GES&0.38$\pm$0.04&0.02$\pm$0.01&0.37$\pm$0.04&0.11$\pm$0.02&0.00$\pm$0.00&0.01$\pm$0.00&0.02$\pm$0.00&0.02$\pm$0.00&0.02$\pm$0.00&0.00$\pm$0.00\\

\rowcolor{Col1}[.22\tabcolsep][2.7mm]GRNBoost&0.09$\pm$0.00&0.00$\pm$0.00&0.08$\pm$0.00&0.02$\pm$0.00&0.01$\pm$0.00&0.87$\pm$0.00&0.95$\pm$0.01&0.88$\pm$0.00&0.89$\pm$0.00&0.65$\pm$0.01\\

\rowcolor{Col1}[.22\tabcolsep][2.7mm]NOTEARS (Linear)&0.88$\pm$0.00&0.12$\pm$0.00&0.88$\pm$0.00&0.38$\pm$0.00&0.00$\pm$0.00&0.00$\pm$0.00&0.00$\pm$0.00&0.00$\pm$0.00&0.00$\pm$0.00&0.00$\pm$0.00\\

\rowcolor{Col1}[.22\tabcolsep][2.7mm]NOTEARS (Linear,L1)&0.88$\pm$0.00&0.12$\pm$0.00&0.88$\pm$0.00&0.38$\pm$0.00&0.00$\pm$0.00&0.00$\pm$0.00&0.00$\pm$0.00&0.00$\pm$0.00&0.00$\pm$0.00&0.00$\pm$0.00\\

\rowcolor{Col1}[.22\tabcolsep][2.7mm]NOTEARS (MLP)&0.16$\pm$0.00&0.01$\pm$0.00&0.14$\pm$0.00&0.03$\pm$0.00&0.01$\pm$0.00&0.60$\pm$0.01&0.80$\pm$0.01&0.61$\pm$0.01&0.61$\pm$0.01&0.25$\pm$0.00\\

\rowcolor{Col1}[.22\tabcolsep][2.7mm]NOTEARS (MLP,L1)&0.98$\pm$0.04&0.15$\pm$0.01&0.98$\pm$0.04&0.59$\pm$0.05&0.00$\pm$0.00&0.00$\pm$0.00&0.00$\pm$0.00&0.00$\pm$0.00&0.00$\pm$0.00&0.00$\pm$0.00\\

\rowcolor{Col1}[.22\tabcolsep][2.7mm]PC&0.30$\pm$0.01&0.02$\pm$0.00&0.28$\pm$0.01&0.08$\pm$0.01&0.01$\pm$0.00&0.02$\pm$0.00&0.03$\pm$0.00&0.02$\pm$0.00&0.03$\pm$0.00&0.00$\pm$0.00\\

\rowcolor{Col1}[.22\tabcolsep][2.7mm]Sortnregress&0.34$\pm$0.00&0.02$\pm$0.00&0.32$\pm$0.00&0.09$\pm$0.00&0.02$\pm$0.00&0.09$\pm$0.00&0.11$\pm$0.00&0.10$\pm$0.00&0.11$\pm$0.00&0.02$\pm$0.00\\
\hline
\rowcolor{Col2}[.22\tabcolsep][2.7mm]GIES&0.30$\pm$0.01&0.02$\pm$0.01&0.29$\pm$0.01&0.08$\pm$0.01&0.01$\pm$0.00&0.02$\pm$0.00&0.03$\pm$0.01&0.03$\pm$0.00&0.03$\pm$0.00&0.00$\pm$0.00\\

\rowcolor{Col2}[.22\tabcolsep][2.7mm]DCDFG-MLP&0.26$\pm$0.08&0.02$\pm$0.01&0.23$\pm$0.08&0.05$\pm$0.01&0.04$\pm$0.01&0.13$\pm$0.10&0.15$\pm$0.09&0.13$\pm$0.10&0.13$\pm$0.10&0.12$\pm$0.09\\

\rowcolor{Col2}[.22\tabcolsep][2.7mm]DCDI-DSF&0.21$\pm$0.00&0.01$\pm$0.00&0.20$\pm$0.00&0.06$\pm$0.00&0.01$\pm$0.00&0.04$\pm$0.00&0.05$\pm$0.00&0.05$\pm$0.00&0.05$\pm$0.00&0.01$\pm$0.00\\

\rowcolor{Col2}[.22\tabcolsep][2.7mm]DCDI-G&0.26$\pm$0.00&0.02$\pm$0.00&0.25$\pm$0.00&0.07$\pm$0.00&0.01$\pm$0.00&0.03$\pm$0.00&0.04$\pm$0.01&0.03$\pm$0.00&0.04$\pm$0.00&0.00$\pm$0.00\\
\hline
Random (k=1000)&0.14$\pm$0.01&0.01$\pm$0.00&0.12$\pm$0.01&0.03$\pm$0.00&0.01$\pm$0.01&0.01$\pm$0.00&0.01$\pm$0.00&0.01$\pm$0.00&0.01$\pm$0.00&0.01$\pm$0.00\\
\bottomrule

\end{tabular}}
\end{adjustbox}
    \label{tab:benchmark_rpe1}
\end{table*}

\begin{table*}[h!]
\centering
\footnotesize
\caption{Performance results on K562 Dataset for the biological evaluation, for each evidence type. Pooled correspond to pooling all the extracted databases into one network for evaluation.}
\begin{adjustbox}{center, max width=11cm}
{\renewcommand{\arraystretch}{1.15}
\begin{tabular}{l|ccccc|ccccc}
\hline
Model&\multicolumn{5}{@{}c@{}|}{Precision}&\multicolumn{5}{@{}c@{}}{Recall}\\
\hline
&Pooled&CORUM&PPI (N)&PPI (P)&CHIP&Pooled&CORUM&PPI (N)&PPI (P)&CHIP\\
\hline

\rowcolor{Col1}[.22\tabcolsep][2.7mm]GES&0.59$\pm$0.02&0.18$\pm$0.01&0.59$\pm$0.02&0.26$\pm$0.01&0.00$\pm$0.00&0.01$\pm$0.00&0.03$\pm$0.00&0.01$\pm$0.00&0.01$\pm$0.00&0.00$\pm$0.00\\

\rowcolor{Col1}[.22\tabcolsep][2.7mm]GRNBoost&0.22$\pm$0.00&0.02$\pm$0.00&0.21$\pm$0.00&0.07$\pm$0.00&0.01$\pm$0.00&0.68$\pm$0.00&0.81$\pm$0.01&0.69$\pm$0.00&0.71$\pm$0.00&0.32$\pm$0.00\\

\rowcolor{Col1}[.22\tabcolsep][2.7mm]NOTEARS (Linear)&0.88$\pm$0.00&0.50$\pm$0.00&0.88$\pm$0.00&0.50$\pm$0.00&0.00$\pm$0.00&0.00$\pm$0.00&0.00$\pm$0.00&0.00$\pm$0.00&0.00$\pm$0.00&0.00$\pm$0.00\\

\rowcolor{Col1}[.22\tabcolsep][2.7mm]NOTEARS (Linear,L1)&0.83$\pm$0.00&0.50$\pm$0.00&0.83$\pm$0.00&0.50$\pm$0.00&0.00$\pm$0.00&0.00$\pm$0.00&0.00$\pm$0.00&0.00$\pm$0.00&0.00$\pm$0.00&0.00$\pm$0.00\\

\rowcolor{Col1}[.22\tabcolsep][2.7mm]NOTEARS (MLP)&0.29$\pm$0.00&0.02$\pm$0.00&0.27$\pm$0.00&0.08$\pm$0.00&0.01$\pm$0.00&0.65$\pm$0.01&0.53$\pm$0.01&0.66$\pm$0.01&0.62$\pm$0.01&0.27$\pm$0.02\\

\rowcolor{Col1}[.22\tabcolsep][2.7mm]NOTEARS (MLP,L1)&0.77$\pm$0.09&0.06$\pm$0.10&0.77$\pm$0.09&0.23$\pm$0.09&0.00$\pm$0.00&0.00$\pm$0.00&0.00$\pm$0.00&0.00$\pm$0.00&0.00$\pm$0.00&0.00$\pm$0.00\\

\rowcolor{Col1}[.22\tabcolsep][2.7mm]PC&0.46$\pm$0.01&0.09$\pm$0.00&0.46$\pm$0.01&0.16$\pm$0.01&0.01$\pm$0.00&0.02$\pm$0.00&0.03$\pm$0.00&0.02$\pm$0.00&0.02$\pm$0.00&0.00$\pm$0.00\\

\rowcolor{Col1}[.22\tabcolsep][2.7mm]Sortnregress&0.63$\pm$0.00&0.21$\pm$0.00&0.63$\pm$0.00&0.29$\pm$0.00&0.00$\pm$0.00&0.07$\pm$0.00&0.25$\pm$0.00&0.07$\pm$0.00&0.11$\pm$0.00&0.01$\pm$0.00\\
\hline
\rowcolor{Col2}[.22\tabcolsep][2.7mm]GIES&0.52$\pm$0.01&0.11$\pm$0.00&0.52$\pm$0.01&0.19$\pm$0.00&0.01$\pm$0.00&0.02$\pm$0.00&0.04$\pm$0.00&0.02$\pm$0.00&0.02$\pm$0.00&0.00$\pm$0.00\\

\rowcolor{Col2}[.22\tabcolsep][2.7mm]DCDFG-MLP&0.40$\pm$0.01&0.04$\pm$0.01&0.40$\pm$0.01&0.11$\pm$0.01&0.00$\pm$0.00&0.07$\pm$0.01&0.07$\pm$0.03&0.07$\pm$0.01&0.07$\pm$0.02&0.00$\pm$0.00\\

\rowcolor{Col2}[.22\tabcolsep][2.7mm]DCDI-DSF&0.39$\pm$0.00&0.07$\pm$0.00&0.38$\pm$0.00&0.14$\pm$0.00&0.00$\pm$0.00&0.03$\pm$0.00&0.06$\pm$0.00&0.03$\pm$0.00&0.04$\pm$0.00&0.00$\pm$0.00\\

\rowcolor{Col2}[.1\tabcolsep][2.7mm]DCDI-G&0.44$\pm$0.02&0.12$\pm$0.00&0.44$\pm$0.02&0.18$\pm$0.00&0.00$\pm$0.00&0.02$\pm$0.00&0.06$\pm$0.00&0.02$\pm$0.00&0.03$\pm$0.00&0.00$\pm$0.00\\
\hline
Random (k=1000)&0.26$\pm$0.02&0.02$\pm$0.00&0.25$\pm$0.02&0.07$\pm$0.01&0.01$\pm$0.00&0.01$\pm$0.00&0.01$\pm$0.00&0.01$\pm$0.00&0.01$\pm$0.00&0.00$\pm$0.00\\
\bottomrule
\end{tabular}}
\end{adjustbox}
    
    \label{tab:benchmark_k562}
\end{table*}

\section{Partition sizes}
\label{app:partition}
We here recapitulate the partition sizes used to be able to run each method in \autoref{tab:partitions}.

\begin{table}[]
    \centering
    \caption{Partition sizes used for each model. $-1$ means that the graph was not partitioned.}
    \begin{tabular}{lc}
    \toprule
        Model name &  partition size\\
        \midrule
         PC & 30\\
         GES & 30\\
         GIES & 30\\
         NOTEARS (Linear) & -1\\
         NOTEARS (Linear, L1) & -1\\
         NOTEARS (MLP) & -1\\
         NOTEARS (MLP, L1) & -1\\
         DCDI-DSF & 50\\
         DCDI-G & 50\\
         DCDFG-MLP & -1 \\
         GRNBoost & -1 \\
         Sortnregress & -1\\
         \bottomrule
    \end{tabular}
    
    \label{tab:partitions}
\end{table}

\section{Model ranking}
\label{app:ranking}

We here present a simple unbiased way of ranking the different models based on the mean Wasserstein distance and the FOR. We separate the rankings per cell type. First, we create a preliminary ranking for each evaluation metrics. 
We first rank by the mean score across the different seeds. Then, we assign the same rank to all models having overlapping confidence interval, starting from the bottom of the ranking. Finally, for each model, we take their average rank across the evaluation specific rankings. This ranking thus gives the same weight to each evaluation method. Results are summarized in \autoref{tab:ranking_k562} for the K562 cell line and in \autoref{tab:ranking_rpe1} for the RPE1 cell line. 

\begin{table}[]
    \centering
    \caption{Complete ranking on the K562 cell line. }
\begin{tabular}{lrrrll}
\toprule
           Model &  Rank  &  Rank  &  Mean Rank & Wasserstein  & FOR \\
           & Wasserstein & FOR & & Distance & \\
\midrule
                   GRNBoost &                             12 &                             1 &      6.500 &           0.133 ± 0.000 &          0.126 ± 0.000 \\
             DCDI-G &                              1 &                            12 &      6.500 &           0.183 ± 0.001 &          0.182 ± 0.021 \\
       Sortnregress &                              2 &                            12 &      7.000 &           0.171 ± 0.000 &          0.184 ± 0.000 \\
           DCDI-DSF &                              5 &                            12 &      8.500 &           0.163 ± 0.003 &          0.181 ± 0.017 \\
                GES &                              5 &                            12 &      8.500 &           0.164 ± 0.006 &          0.185 ± 0.025 \\
   NOTEARS (Linear) &                              5 &                            12 &      8.500 &           0.164 ± 0.000 &          0.188 ± 0.027 \\
               GIES &                              7 &                            12 &      9.500 &           0.155 ± 0.004 &          0.183 ± 0.025 \\
                 PC &                              7 &                            12 &      9.500 &           0.152 ± 0.002 &          0.187 ± 0.025 \\
      NOTEARS (MLP) &                             12 &                            12 &     12.000 &           0.139 ± 0.000 &          0.179 ± 0.003 \\
          DCDFG-MLP &                             12 &                            12 &     12.000 &           0.143 ± 0.004 &          0.180 ± 0.000 \\
NOTEARS (Linear,L1) &                             12 &                            12 &     12.000 &           0.140 ± 0.000 &          0.188 ± 0.027 \\
   NOTEARS (MLP,L1) &                             12 &                            12 &     12.000 &           0.113 ± 0.032 &          0.188 ± 0.027 \\

\bottomrule
\end{tabular}
\label{tab:ranking_k562}
\end{table}

\begin{table}[]
    \centering
    \caption{Complete ranking on the RPE1 cell line. }
\begin{tabular}{lrrrll}
\toprule
           Model &  Rank  &  Rank  &  Mean Rank & Wasserstein  & FOR \\
           & Wasserstein & FOR & & Distance & \\
\midrule
           Sortnregress &                              2 &                             3 &      2.500 &           0.170 ± 0.000 &          0.114 ± 0.000 \\
           GRNBoost &                             12 &                             1 &      6.500 &           0.110 ± 0.000 &          0.110 ± 0.000 \\
             DCDI-G &                              1 &                            12 &      6.500 &           0.188 ± 0.005 &          0.135 ± 0.007 \\
      NOTEARS (MLP) &                             11 &                             3 &      7.000 &           0.138 ± 0.000 &          0.111 ± 0.012 \\
                 PC &                              7 &                            12 &      9.500 &           0.164 ± 0.004 &          0.131 ± 0.020 \\
           DCDI-DSF &                              7 &                            12 &      9.500 &           0.166 ± 0.003 &          0.133 ± 0.007 \\
   NOTEARS (MLP,L1) &                              7 &                            12 &      9.500 &           0.166 ± 0.004 &          0.138 ± 0.016 \\
   NOTEARS (Linear) &                              7 &                            12 &      9.500 &           0.164 ± 0.000 &          0.138 ± 0.016 \\
NOTEARS (Linear,L1) &                              7 &                            12 &      9.500 &           0.164 ± 0.000 &          0.138 ± 0.016 \\
          DCDFG-MLP &                             11 &                            12 &     11.500 &           0.132 ± 0.018 &          0.129 ± 0.009 \\
                GES &                             11 &                            12 &     11.500 &           0.151 ± 0.012 &          0.133 ± 0.017 \\
               GIES &                             11 &                            12 &     11.500 &           0.143 ± 0.004 &          0.140 ± 0.008 \\
\bottomrule
\end{tabular}  
    \label{tab:ranking_rpe1}
\end{table}

\section{Run time} 
\label{app:runtime}
We here present the average run time for each method in \autoref{tab:run_k562} for the K562 cell line and in \autoref{tab:run_rpe1} for the RPE1 cell line

\begin{table}[]
    \centering
    \caption{Run time in wall clock hours for each method on the K562 cell line. }
    \begin{tabular}{lr}
\toprule
              Model & Run time (hours) \\
\midrule
           GRNBoost &  0.087988 \\
       Sortnregress &  0.089441 \\
               GIES &  0.653034 \\
   NOTEARS (MLP,L1) &  1.526646 \\
           DCDI-DSF &  1.586062 \\
          DCDFG-MLP &  1.768952 \\
                 PC &  2.610882 \\
                GES &  3.337906 \\
             DCDI-G &  4.600406 \\
NOTEARS (Linear,L1) &  5.158183 \\
   NOTEARS (Linear) &  5.814321 \\
      NOTEARS (MLP) &  9.134294 \\
\bottomrule
\end{tabular}
    
    \label{tab:run_k562}
\end{table}

\begin{table}[]
    \centering
    \caption{Run time in wall clock hours for each method on the RPE1 cell line. }
    \begin{tabular}{lr}
\toprule
              Model & Run time (hours) \\
\midrule
       Sortnregress &  0.045235 \\
           GRNBoost &  0.058020 \\
           DCDI-DSF &  0.494506 \\
          DCDFG-MLP &  1.029798 \\
               GIES &  1.082506 \\
   NOTEARS (MLP,L1) &  1.116924 \\
                GES &  1.333881 \\
             DCDI-G &  1.793924 \\
   NOTEARS (Linear) &  1.807665 \\
NOTEARS (Linear,L1) &  1.821760 \\
                 PC &  5.610735 \\
      NOTEARS (MLP) &  8.455514 \\
\bottomrule
\end{tabular}
    
    \label{tab:run_rpe1}
\end{table}

\end{document}